\documentclass[acmtog]{acmart}

\usepackage[utf8]{inputenc}
\usepackage{amsmath}

\AtBeginDocument{%
 \providecommand\BibTeX{{%
  \normalfont B\kern-0.5em{\scshape i\kern-0.25em b}\kern-0.8em\TeX}}}

\setcopyright{acmlicensed}
\acmJournal{TOG}
\acmYear{2022}
\acmVolume{41}
\acmNumber{6}
\acmArticle{220}
\acmMonth{12}
\acmDOI{10.1145/3550454.3555491}

\acmSubmissionID{370}

\setcitestyle{square}

\begin{document}

\title{Neural Cloth Simulation}

\author{Hugo Bertiche}
\email{hugo\_bertiche@hotmail.com}
\orcid{0000-0002-6632-1902}
\author{Meysam Madadi}
\email{mmadadi@cvc.uab.cat}
\orcid{0000-0002-7384-5712}
\author{Sergio Escalera}
\email{sescalera@ub.edu}
\orcid{0000-0003-0617-8873}
\affiliation{%
 \institution{Universitat de Barcelona}
 \streetaddress{Gran Via de les Corts Catalanes, 585}
 \city{Barcelona}
 \country{Spain}
 \postcode{08005}
}
\affiliation{
 \institution{Computer Vision Center, UAB}
 \streetaddress{Campus UAB Edifici O}
 \city{Bellaterra}
 \country{Spain}
 \postcode{08193}
}


\renewcommand{\shortauthors}{Bertiche et al.}

\begin{abstract}
 We present a general framework for the garment animation problem through unsupervised deep learning inspired in physically based simulation. Existing trends in the literature already explore this possibility. Nonetheless, these approaches do not handle cloth dynamics. Here, we propose the first methodology able to learn realistic cloth dynamics unsupervisedly, and henceforth, a general formulation for neural cloth simulation. The key to achieve this is to adapt an existing optimization scheme for motion from simulation based methodologies to deep learning. Then, analyzing the nature of the problem, we devise an architecture able to automatically disentangle static and dynamic cloth subspaces by design. We will show how this improves model performance. Additionally, this opens the possibility of a novel motion augmentation technique that greatly improves generalization. Finally, we show it also allows to control the level of motion in the predictions. This is a useful, never seen before, tool for artists. We provide of detailed analysis of the problem to establish the bases of neural cloth simulation and guide future research into the specifics of this domain.
\end{abstract}



\begin{teaserfigure}
  \includegraphics[width=\textwidth]{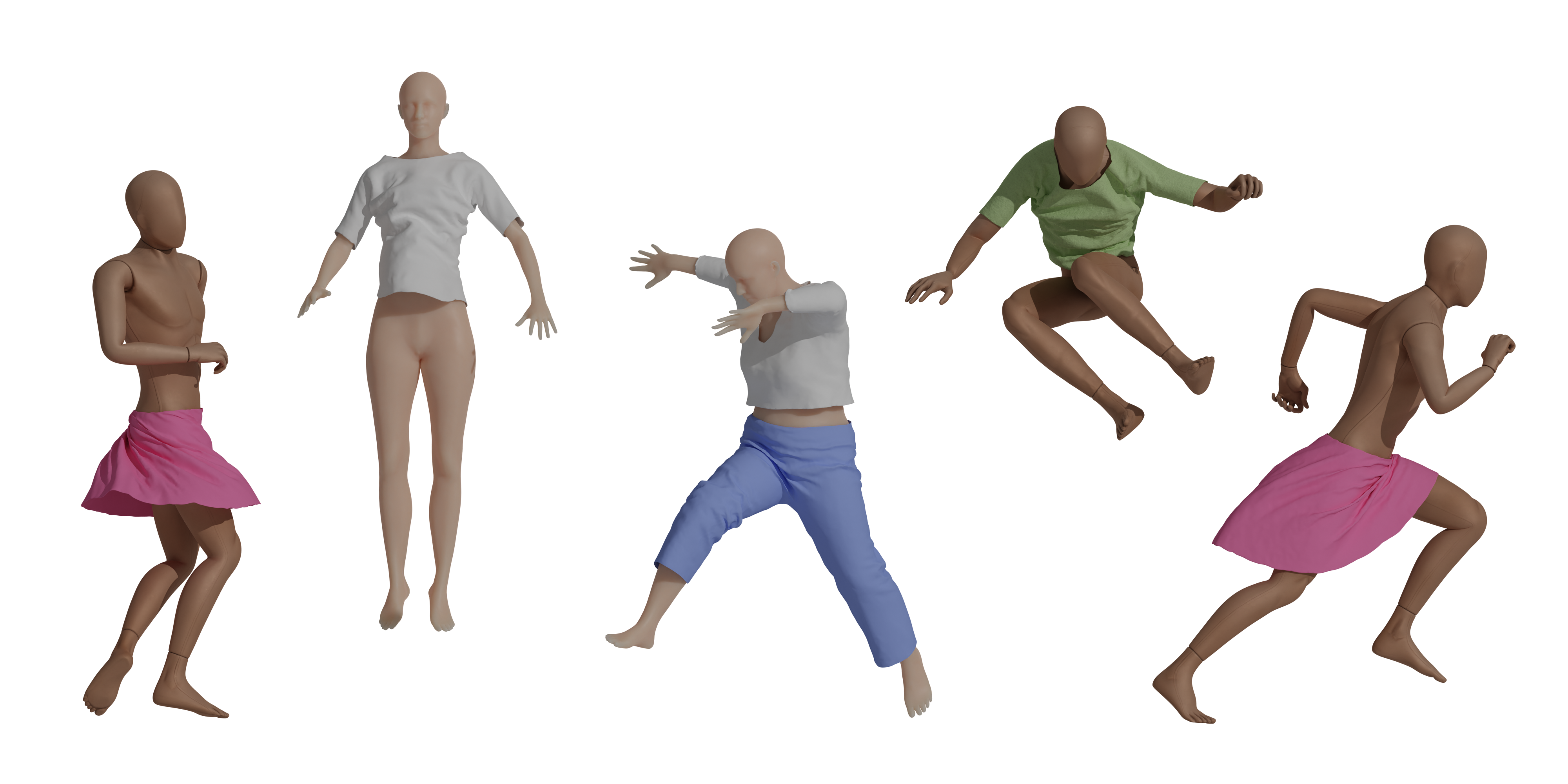}
  \caption{We present a general framework for the garment animation problem through neural cloth simulation. More specifically, an unsupervised deep learning methodology inspired in physically based simulation. Ours is the first methodology able to learn cloth dynamics without any ground truth data.}
  \label{fig:teaser}
\end{teaserfigure}

\maketitle

\section{Introduction}

Cloth animation has been a focus of research during decades. Mainly due to its numerous applications in the entertainment and fashion industry. Firstly, computer graphics approaches relied on physically based simulation to animate cloth \cite{baraff1998large, muller2007position, narain2012adaptive, liu2013fast, pfaff2014adaptive, hahn2014subspace, macklin2016xpbd}. While it is possible to obtain physically accurate results, these methodologies are computationally expensive, which makes them unsuitable for scenarios where real-time is a requirement, such as video-games or VR/AR. The research community, inspired by the success of deep learning in other 3D tasks \cite{socher2012convolutional, richardson20163d, qi2017pointnet, han2017deepsketch2face, arsalan2017synthesizing, omran2018neural, madadi2020smplr} and its fast inference properties, has recently shown interest in neural networks as a suitable alternative for fast garment animation.

Initially, authors proposed supervised solutions \cite{wang2019learning, gundogdu2019garnet, patel2020tailornet, bertiche2020cloth3d, bertiche2021deepsd}. To simplify the problem, garments are usually skinned w.r.t. the underlying skeleton that drives the body motion. Then, the network task is to predict cloth deformations in rest pose. These approaches present some drawbacks. Supervised learning requires huge volumes of computationally expensive data. Moreover, this process has to be repeated for each garment and body. Also, more often than not, data requires heavy pre-processing to be ready for training. Finally, predictions usually present body penetrations, which motivates the use of post-processing or strong regularization terms. Authors of \cite{10.1145/3478513.3480479, santesteban2022snug} identified these drawbacks and proposed unsupervised learning schemes. While these approaches addressed some of the drawbacks of supervised methodologies, they do not handle cloth dynamics. On one hand, PBNS \cite{10.1145/3478513.3480479} uses a static formulation. Then, it is unable to learn cloth dynamics. On the other hand, the authors of SNUG \cite{santesteban2022snug} propose to use an inertia term from the computer graphics literature. Nonetheless, their adaptation of the inertia term to deep learning temporally smooths cloth particle velocities (minimizes accelerations). We will see this is not the same as learning cloth dynamics (Sec.~\ref{sec:vs_sota}).

We present the first methodology able to learn real cloth dynamics without the need of ground truth data. By doing so, we define the first general framework for neural garment simulation. The list of our contributions is as follows:
\begin{enumerate}
  \item \textbf{Unsupervised Cloth Dynamics.} In the computer graphics literature we can find simulation based works that recast the equations of motion as an optimization problem. This means that a similar solution can be applied to deep learning. We adapt this solution to unsupervised training of neural networks. We will show how our methodology is the first to be able to learn cloth dynamics in an unsupervised fashion.
  \item \textbf{Disentangled Cloth Subspace.} We analyze the nature of the garment animation problem to motivate a novel architecture that allows an automatic disentanglement of cloth static and dynamic deformations at a subspace level. We will see how this improves model performance as well as allowing control over cloth dynamics. Additionally, we leverage the disentanglement to propose a novel motion augmentation technique that further improves model generalization.
  \item \textbf{In-depth Analysis on Neural Garment Simulation.} Unsupervised garment animation differs from other deep learning tasks, supervised or unsupervised. We provide of detailed analysis on the problem to understand its peculiarities and help to establish the bases of neural simulation for garments.
\end{enumerate}
The rest of the paper is as follows. In Sec.~\ref{sec:sota} we review the literature on cloth simulation and deep learning based methodologies on garments. Next, Sec.~\ref{sec:methodology} describes the methodology we propose. Then, Sec.~\ref{sec:results} contains an analysis on the different metrics, an ablation study and a comparison with the state-of-the-art. Finally, in Sec.~\ref{sec:conclusions} we discuss limitations and future research.

\section{Related Work}
\label{sec:sota}

\paragraph{Cloth simulation.} Computer graphics has been tackling the cloth animation problem for decades. The first advances in the field were done by \cite{weil1986synthesis}, as static geometry based models. Later, researchers developed elastic continuum models for cloth \cite{feynman1986modeling, terzopoulos1987elastically, baraff1998large, carignan1992dressing} that permit dynamic simulations. On the other hand, other authors \cite{haumann1987modeling, breen1992physically, provot1995deformation} noted cloth is not a continuum, but a combination of mechanical interactions between cloth yarns. From here, alternative particle based formulations for cloth were developed, like the mass-spring model. Later, the work of \cite{baraff1998large} presented triangle-based formulation for cloth that allowed fast simulation of complex garments. To this day, this work is still the foundation of many current methodologies for cloth simulation \cite{narain2012adaptive, pfaff2014adaptive}. Later, \cite{kaldor2008simulating, kaldor2010efficient} proposed modelling cloth at yarn-level to achieve highly realistic behaviour. These methodologies --while accurate-- are computationally expensive, therefore, not suitable for many applications that demand real-time performance. Simpler and more efficient formulations have been developed in favor of a faster simulation \cite{muller2007position, liu2013fast, macklin2016xpbd} at the cost of accuracy or realism. Nonetheless, realistic simulation of fine cloth dynamics in real-time is still unfeasible with standard simulation. Specially as the number of cloth triangles increase while also considering that some of these real-time applications often require computational resources for other tasks. Subspace physics has proved a valid fast alternative for soft body simulation \cite{teng2014simulating, pan2015subspace}, and recently in combination with learnt representations \cite{Fulton:LSD:2018, 10.1145/3450626.3459754}. This alternative has also been proposed for clothing \cite{de2010stable, kim2013near, hahn2014subspace}, although in practice, collisions are not properly handled. Then, while computer graphics offers realistic and accurate solutions, efficient real-time cloth animation remains an open challenge.

\paragraph{Deep learning.} During recent years, neural networks have proved their usefulness in many complex tasks. One of their main advantages is a fast inference time. Then, given their success in challenging 3D problems \cite{socher2012convolutional, richardson20163d, qi2017pointnet, han2017deepsketch2face, arsalan2017synthesizing, omran2018neural, madadi2020smplr}, researchers have already turned to deep-based solutions for garment animation. Most of the current literature on the domain relies on supervised learning \cite{gundogdu2019garnet, wang2019learning, patel2020tailornet, bertiche2020cloth3d, bertiche2021deepsd, santesteban2021self, pfaff2020learning, santesteban2019learning, zhang2021dynamic}. To this end, it is necessary to run hundreds or thousands of offline physics based simulations to gather the data required for training. Data gathering needs to be repeated for every garment, body and fabric parameters. This hurts the scalability of supervised solutions. Additionally, supervised learning is biased towards lower frequencies, yielding overly smooth garments. Moreover, supervision does not guarantee physical constraints are satisfied. Finally, simulators display a chaotic behaviour. Garment simulation on top of very similar body motions may result in considerably different outputs. In practice, this means noisy data, which hinders training. Authors of \cite{10.1145/3478513.3480479} proposed PBNS, an alternative unsupervised solution that does not suffer from the drawbacks related to simulated data. To do so, they propose formulating physically based constraints as energy losses. This permits learning garment deformations without ground truth data. Nonetheless, their formulation is purely static. They do not handle cloth dynamics. Similarly, SNUG \cite{santesteban2022snug} uses the same unsupervised scheme as PBNS to learn garment deformations. With the only addition of an inertia loss term from the physics based simulation literature \cite{liu2013fast, martin2011example, gast2015optimization} to model dynamic deformations. However, in this work we will see how their adaptation of the inertia term is not modelling true cloth dynamics. Following this trend of unsupervised learning for garment animation, we present the first work able to learn cloth dynamics. Thus, defining the first general framework for neural cloth simulation. We also propose a novel model architecture that automatically disentangles static and dynamic cloth deformations. This, in turn, shows improved performance and interesting novel properties.

\section{Methodology}
\label{sec:methodology}

The goal of this work is to define a deep-learning based methodology for the garment animation problem. This problem corresponds to the animation of cloth draped around skinned 3D body models. Following the current trend \cite{10.1145/3478513.3480479, santesteban2022snug}, we propose an unsupervised training inspired on physically based simulation. We additionally achieve a disentanglement of static and dynamic cloth deformations by considering the nature of both cases in our model design.

\subsection{Neural Cloth Subspace Solver}

\paragraph{Defining the problem.} Our methodology for learning cloth dynamics unsupervisedly is inspired in classical computer graphics physical simulation \cite{baraff1998large, muller2007position, liu2013fast}. To this end, cloth is presented as a particle system $\mathbf{x}\in\mathbb{R}^{N\times3}$. Cloth solvers compute the cloth configuration at instant $t$ from the previous cloth state, which is defined by the particle locations and velocities at $t - 1$. Additionally, the solver must also consider external forces --colliders, wind, etc-- that act on the cloth. Within the scope of this work --garment animation-- the external forces correspond to gravity, which is constant, and the collisions with the underlying skinned 3D model draped with the clothes. Then, given a skinned 3D model parameterized by $\boldsymbol\theta$ (pose and location in our experiments, but potentially any parameterization, like body shape \cite{loper2015smpl}), the solver can be written as:
\begin{equation}
  \mathbf{x}_t = f(\boldsymbol {\theta}_t, \mathbf{x}_{t-1}, \mathbf{v}_{t-1}),
  \label{eq:integrator}
\end{equation}
where $\mathbf{v}$ is the particle velocities. Velocities will depend on the current and previous particle locations, therefore, we can rewrite the expression as $\mathbf{x}_t = f(\boldsymbol\theta_t, \mathbf{x}_{t-1}, \mathbf{x}_{t-2})$. Likewise, we have that $\mathbf{x}_{t-1} = f(\boldsymbol\theta_{t-1}, \mathbf{x}_{t-2}, \mathbf{x}_{t-3})$. The same could be done for $\mathbf{x}_{t-2}$, $\mathbf{x}_{t-3}$ and so on and so forth. This yields the following:
\begin{equation}
    \mathbf{x}_t = \hat{f}(\boldsymbol\theta_t, \boldsymbol\theta_{t-1}, \boldsymbol\theta_{t-2}, ..., \boldsymbol\theta_0, \mathbf{x}_0).
    \label{eq:well_posed}
\end{equation}
Thus, assuming the 3D body model parameterized by $\boldsymbol\theta$ and the gravity are the only external forces, the cloth can be fully parameterized by the body pose history --or motion-- $\boldsymbol\Theta_t = \{\boldsymbol\theta_t, \boldsymbol\theta_{t-1}, \boldsymbol\theta_{t-2}, ..., \boldsymbol\theta_0\}$. 
We can intuitively assume that early poses will have less impact on the current cloth state.
This allows to safely discard poses outside a given temporal window. To ensure the problem remains well-posed, the temporal window size should be sufficiently large. Garments that may show more complex, longer dynamics --like dresses or skirts-- will require a larger temporal window size than garments that will not show complex dynamics --like tight pants.
We also identify separately the static case, a special case of garment animation in which there is no body or cloth motion. This corresponds to the result of draping a body staying still in a given pose during an \textit{infinite} --long enough-- amount of time. For such case, we have $\mathbf{x}_t = \mathbf{x}_{t-1}=...=\mathbf{x}_0$ and $\boldsymbol\theta_t = \boldsymbol\theta_{t-1} = ... = \boldsymbol\theta_0$, hence, the cloth can --and must-- be fully parameterized using only $\boldsymbol\theta_t$. 

\paragraph{Cloth subspace.} Once we have been able to establish a relationship between the body motion space and cloth space, we are implicitly declaring that a clothing subspace $\mathcal{Z}\subset\mathbb{R}^d$ exists, with $d << N\times3$. That is, $\boldsymbol\Theta_t \rightarrow \mathbf{z}_t \rightarrow \mathbf{x}_t$, with $\mathbf{z}_t\in\mathcal{Z}$. Then, inspired on subspace physics \cite{de2010stable, kim2013near, hahn2014subspace}, our proposed model must be able to solve the next cloth configuration from the current subspace encoding:
\begin{equation}
  \mathbf{z}_{t+1} = g(\boldsymbol\theta_{t+1}, \mathbf{z}_t),
\end{equation}
where $g(\cdot)$ corresponds to a neural cloth subspace solver. Additionally, it must be designed in a way that ensures that given a static sample the subspace encoding does not change, $\mathbf{z}_t = \mathbf{z}_{t-1} = ... = \mathbf{z}_0$. The optimal solution must naturally fall back to a static formulation when there is no input body motion. We know the static solution is a special case of the general problem, likewise, the subspace of all static cloth states $\mathbf{z}^S\in\mathcal{Z}^S$ is a subspace of the subspace of all possible cloth states, $\mathcal{Z}^S\subset\mathcal{Z}$, in the same way that the body pose space is a subspace of the body motion space. That is:
\begin{equation}
  \boldsymbol\theta_t \rightarrow \mathbf{z}^S_t, \quad
  \boldsymbol\Theta_t \rightarrow \mathbf{z}_t.
  \label{eq:pose_motion_manifold}
\end{equation}
We know the pose space is a single continuous manifold, henceforth, due to eq.~\ref{eq:pose_motion_manifold}, the static cloth subspace must also be a single continuous manifold. Similarly, we can extend this reasoning to the motion space and the full cloth subspace. We can then conclude that subspace $\mathcal{Z}$ is a higher-dimensional manifold around the static subspace $\mathcal{Z}^S$.

\paragraph{Neural network.} Motivated by all of the above, we propose an encoder-decoder recurrent neural network as a suitable architecture for this problem. The model takes body motion $\boldsymbol\Theta_t$ as input, encodes it as $\mathbf{z}_t$ and finally decodes it into the predicted cloth $\mathbf{x}_t$. 
Also, we present a novel disentangled encoder that will enforce by design the expected static and dynamic behaviour. We will show how this results in an improved performance and explainability. Moreover, by disentangling static and dynamic deformations, we allow control over the level of motion in our predictions. This property will help artists to achieve the desired looks for a given application. We also define a novel motion augmentation technique that greatly increases model robustness.
As it is usual in the literature \cite{gundogdu2019garnet, bertiche2020cloth3d, bertiche2021deepsd, patel2020tailornet, 10.1145/3478513.3480479, santesteban2022snug}, the network predicts cloth deformations in rest pose. Then, the garment is skinned w.r.t. the underlying body skeleton and it is posed along with the body mesh.

\subsection{Body Motion Descriptors}
\label{sec:descriptors}

As explained, to fully parameterize the garment state $\mathbf{x}_t$ we need the body pose history, to which we refer as body motion $\boldsymbol\Theta_t$. To keep the problem tractable, we safely truncate the pose history using a reasonable temporal window size. The window size will be directly related to the maximum cloth motion length that our network will be able to learn. Looser garments --like skirts-- require longer temporal windows. For the rest of the paper, we will refer to the truncated pose history as $\boldsymbol\Theta_t = \{\boldsymbol\theta_t, \boldsymbol\theta_{t-1}, \boldsymbol\theta_{t-2}, ..., \boldsymbol\theta_{t-n}\}$. Then, during training, we predict each sample $\mathbf{x}_t$ using $\boldsymbol\Theta_t$ as network input. For inference, the model can be fed with indefinitely long sequences, as new poses $\boldsymbol\theta$ are used to update the hidden recurrent state. 

The naive baseline solution would be to feed body pose sequences --as joint orientations-- with global body velocities. While this would suffice to avoid an ill-posed problem, this representation is sub-optimal and dynamics are entangled with static information. The model would need to learn by itself to extract body motion information from the input poses. This increases the required training data and time, as well as model capacity, while hurting generalization. We propose a set of disentangled descriptors that are more suitable for this problem.

\paragraph{Static descriptors.} To describe the body pose it is common to use joint relative orientations (relative to the parent joint). The usual axis angle or quaternion representations suffer from a many-to-one problem and discontinuities in the rotation space. Thus, we opt for the 6D descriptors proposed in \cite{zhou2019continuity}. We observe relative orientations are local descriptors. While garment deformations depend on the global body configuration. Small changes in the first joints of the kinematic tree can lead to significantly different garment states. Therefore, samples close to each other in the input space would need to be mapped to points very far from each other in the output space, which makes training and generalization more challenging. On the other hand, using global joint orientations would make the input space extremely large and noisy. Rotations around the gravity axis would create completely different inputs, while the output should remain the same. We propose using for each joint --besides the local 6D descriptor-- a unit vector with the \textit{unposed} direction of the gravity. That is:
\begin{equation}
  \hat{\mathbf{g}}_j = \mathbf{R}_j^{-1}\mathbf{g} / g,
  \label{eq:gravity_unposed}
\end{equation}
where $j$ is the joint index, $\mathbf{R}$ is the rotation matrix corresponding to the global joint orientation, $\mathbf{g}$ is the gravity vector and $g = 9.81\text{m/s}^2$. This descriptor contains information about the global orientation of each joint and it is invariant to rotations around the gravity axis. Additionally, it will be correlated with the direction of local cloth deformations --in rest pose-- due to gravity. We concatenate our local and global descriptors, yielding a $9$-dimensional feature array per joint, that is, $\boldsymbol\theta^S\in\mathbb{R}^{K\times9}$ where $K$ is the number of joints.

\paragraph{Dynamic descriptors.} To describe body motion we take the derivatives in time of the joint orientations and locations. Orientation derivatives are computed from the static descriptors explained in the previous paragraph. Then, location derivatives are computed from the joint locations in space. Note that these derivatives would suffer from the same issues as the global joint orientations, i.e. a large and noisy input space. We address the issue by \textit{unposing} derivatives as in eq.~\ref{eq:gravity_unposed} without normalizing them. This greatly reduces the input space as well as defining a descriptor that it is more strongly correlated to the local cloth dynamic deformations due to motion, both in magnitude and direction. We assume no air resistance, therefore, dynamic cloth deformations will appear only when the body is under acceleration. Hence, we use as motion descriptors the first derivative of joint orientations --any rotation implies an acceleration-- and the second derivative of the joint locations. Both descriptors are concatenated into a $12$-dimensional descriptor per joint, which gives us $\boldsymbol\theta^D\in\mathbb{R}^{K\times12}$ ($K$ joints, $9$ dimensions from the first derivative of the static descriptors and $3$ additional dimensions for the joints accelerations in local space). 

Skinned 3D models often have many joints in hands, feet and face which are unlikely to be relevant for garment dynamics. We remove these joints from the input.

\subsection{Model}

\begin{figure*}
  \centering
  \includegraphics[width=\textwidth]{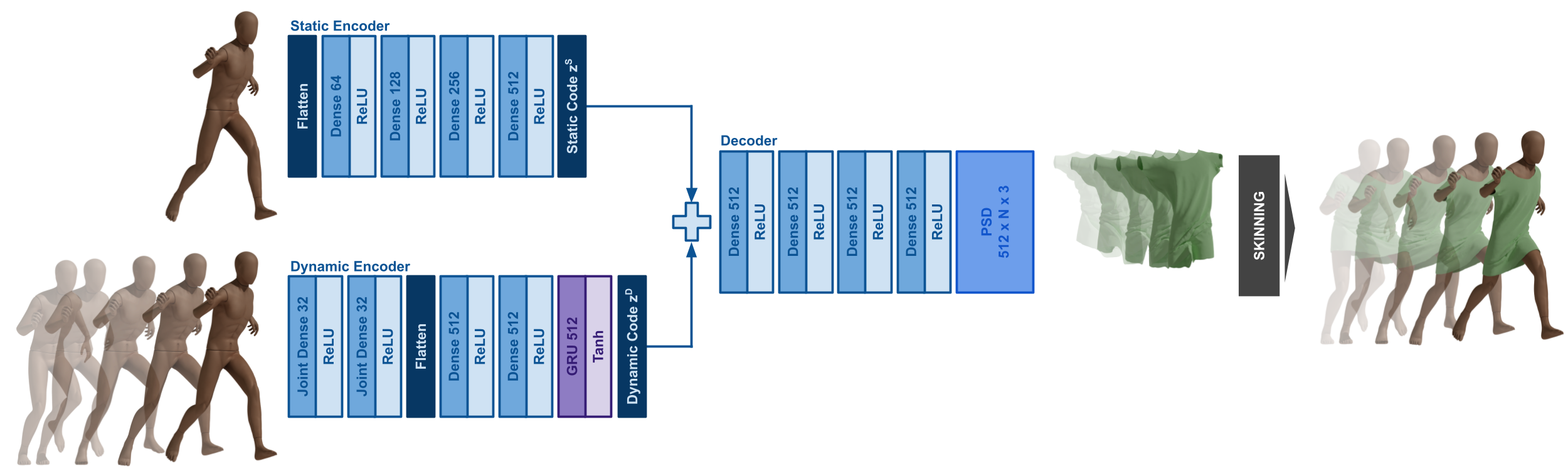}
  \caption{Model architecture. We design our model as a recurrent encoder-decoder. The input of the model is the body motion as described in Sec.~\ref{sec:descriptors}. The encoder is disentangled into a static a dynamic encoder, each fed with the corresponding descriptors. Dynamic encoder layers have no bias, ensuring a direct correlation between input motion and dynamic activations. Encodings are combined by addition and decoded into local cloth deformations. Finally, the garment is skinned with the body.}
  \label{fig:model}
\end{figure*}

In this section we present our model architecture. As explained, we propose a recurrent encoder-decoder network architecture. Our encoder is composed of two different modules, a static and a dynamic encoder, each fed with the corresponding descriptors. Both encodings are combined by addition to be later decoded into local cloth deformations. Finally, the garment is posed along with the body. Fig.~\ref{fig:model} depicts our model.

\paragraph{Static encoder.} We implement this encoder as a set of 4 fully connected layers. The encoder is fed only with the current pose $\boldsymbol\theta^S_t$, which is flattened first into a $9K$-dimensional array. The output of the encoder is a static latent code $\mathbf{z}^S_t$.

\paragraph{Dynamic encoder.} This module is implemented in two blocks. A set of fully connected layers and a Gated Recurrent Unit (GRU). First, 2 fully connected layers are applied to per-joint descriptors individually --as if joints were samples-- to obtain a high-level feature array per joint. We empirically observed this to be beneficial since dynamic descriptors are a concatenation of different modalities. Later, the array is flattened and fed to an additional 2 fully connected layers. Finally, the output is passed through the GRU, which combines it with its hidden state --that encodes the history of dynamics-- to obtain the dynamic latent code $\mathbf{z}^D_t$. Note that $\mathbf{z}^D_t$ is then computed with the whole motion $\boldsymbol\Theta_t$. We observe adding multiple GRUs makes training unstable and hurts model inference speed. All layers of the dynamic encoder have no bias. Without bias, zero input translates to zero output (this is not necessarily true the other way around). This ensures that a static sample will have a null $\mathbf{z}^D_t$, since time derivatives --dynamic descriptors-- will be zero. Thus, addition with $\mathbf{z}^S_t$ will have no impact. Furthermore, samples with high body motion will generally produce high values for $\mathbf{z}^D_t$, creating high dynamic deformations due to a high perturbation of $\mathbf{z}^S_t$. Additionally, the hidden state of the GRU will fade away to zero as long as a constant pose (no motion) is being fed to the model. This means that the latent code $\mathbf{z}$ will naturally fall back to $\mathbf{z}^S$ when motion stops. This also guarantees that, if no motion is present, neither $\mathbf{z}$ nor the output will change, as it will depend only on $\boldsymbol\theta_t$. It would not be possible to ensure this with an entangled encoder or biases in the dynamic branch. Finally, this design allows padding sequences with still frames without altering the value of $\mathbf{z}^D_t$. This property is convenient during training.

Separate encoders have an additional advantage. Their respective input spaces have less variability. Because of this, their latent codes will be more meaningful and they will generalize better. An entangled encoder would need to learn an input space with combinations of static and dynamic descriptors. This would make the task more challenging.

\subsection{Training}
\label{sec:training}

We train our model unsupervisedly by applying losses inspired in physically based simulation, following the trend of \cite{10.1145/3478513.3480479, santesteban2022snug}. Losses are implemented as energy functions of the physical system composed by cloth and body. As training progresses, the network learns to predict garment states that satisfy the energy constraints.

\paragraph{Cloth Model.} In the computer graphics literature we can find multiple ways of defining cloth models. From a simple mass-spring model to more sophisticated continuous approaches that compute triangle deformation energies \cite{baraff1998large, liu2013fast, narain2012adaptive}. To be able to use a cloth formulation within a deep learning framework, the only requirement is that it is differentiable. This makes our approach compatible with most cloth models, allowing them to be freely interchanged. For this work, we implemented mass-spring model as in \cite{10.1145/3478513.3480479}, a squared version of the continuum formulation of \cite{baraff1998large} and Saint Venant Kirchhoff elastic material model as in \cite{santesteban2022snug}. As a loss $\mathcal{L}_\text{cloth}$, this term will penalize in-plane deformations.

\paragraph{Bending Loss.} We implement our bending term for out-of-plane deformations as the squared difference of the angle between adjacent faces w.r.t. the angle in the rest garment \cite{pfaff2014adaptive}. Then, for each pair of adjacent faces:
\begin{equation}
  \mathcal{L}_\text{bending} = k_b\frac{l^2}{8a}(\phi_t - \phi^R)^2,
\end{equation}
where $k_b$ is the bending stiffness, $l$ is the length of the common edge, $a$ is the summation of the area of both faces, $\phi_t$ is the dihedral angle and $\phi^R$ is the dihedral angle in the rest garment. With this formulation, the garment will try to retain its original shape. Scaling the loss as a function of the edge length and triangles area makes it agnostic to mesh resolution and connectivity.

\paragraph{Collisions.} In computer graphics simulations, cloth interaction with external objects is obtained by detection and solving of collisions. Similarly, we implement a loss term that penalizes collisions and creates repelling gradients, pushing cloth vertices outside the body \cite{bertiche2021deepsd, 10.1145/3478513.3480479}:
\begin{equation}
  \mathcal{L}_\text{collision} = k_c\text{min}(\text{d}(\mathbf{x}_t; \boldsymbol\theta_t) - \epsilon, 0)^2,
\end{equation}
where $k_c$ is a balancing factor, $\text{d}(\cdot; \boldsymbol\theta)$ is the signed distance to a body mesh parameterized by $\boldsymbol\theta$, with negative values inside, and $\epsilon$ is a small threshold to ensure robustness. 

\paragraph{Inertia Loss.} Following the laws of motion, a moving object will retain its velocity unless forces act on it. Thus, differences in the location of the cloth particles $\mathbf{x}_t$ and the projected location obtained with the previous velocity $\mathbf{x}^\text{proj}_t = \mathbf{x}_{t-1} + \mathbf{v}_{t-1}\Delta t = 2\mathbf{x}_{t-1} - \mathbf{x}_{t-2}$ are due to other acting forces. A similar observation has already been made in the context of simulation \cite{liu2013fast, martin2011example, gast2015optimization}. This led to the possibility of obtaining the next garment state by finding the critical points of the following expression:
\begin{equation}
  h(\mathbf{x}_t) = \frac{1}{2}(\mathbf{x}_t - \mathbf{x}^\text{proj}_t)^T\mathbf{M}(\mathbf{x}_t - \mathbf{x}^\text{proj}_t) + \Delta t^2E(\mathbf{x}_t),
  \label{eq:motion_optimization}
\end{equation}
where $\mathbf{M}$ is the mass matrix of the particle system, $\Delta t$ is the time step of the simulation and $E(\cdot)$ is the potential representing internal and external forces. This, similar to \cite{santesteban2022snug}, leads to the following loss term for each particle:
\begin{equation}
  \mathcal{L}_\text{inertia} = \frac{1}{2\Delta t^2}m(x_t - x^\text{proj}_t)^2,
\end{equation}
where $m$ is the particle mass. Since $\mathbf{x}^\text{proj}_t$ depends on $\mathbf{x}_{t-1}$ and $\mathbf{x}_{t-2}$, we need to run the model for $\boldsymbol\Theta_{t-1}$ and $\boldsymbol\Theta_{t-2}$ as well. It is crucial not to back-propagate gradients through $\mathbf{x}_{t-1}$ and $\mathbf{x}_{t-2}$. Otherwise, while the loss value gets lower, the model will not show true cloth dynamics. The reason for this is that $\mathbf{x}_t$ would have influence in the location of $\mathbf{x}_{t-1}$ and $\mathbf{x}_{t-2}$ by generating pulling or pushing gradients. Thus, information would be travelling back in time. Note how simulation based related works from which we extract this loss term do not optimize $\mathbf{x}_{t-1}$ or $\mathbf{x}_{t-2}$ to satisfy eq.~\ref{eq:motion_optimization}. One important observation is that this loss will penalize differences in velocities. Thus, whenever the underlying body skeleton joints present no rotations or accelerations, there will not be accelerations in the cloth --since it is attached by blend weights to the skeleton-- and gradients from this term will be zero. Then, no dynamic deformations would be necessary to satisfy the loss. This is consistent with the explanation in Sec.~\ref{sec:descriptors} regarding the choice of dynamic descriptors.

\paragraph{Gravity.} As previous unsupervised approaches, we include the effects of gravity by implementing its potential energy as a loss:
\begin{equation}
  \mathcal{L}_\text{gravity} = -\mathbf{M}\mathbf{x}_t\mathbf{g}.
\end{equation}
This term will push vertices in the direction of the gravity, weighted by particles mass and gravity.

\section{Results}
\label{sec:results}

In this section we explain the results obtained with the proposed methodology. First, we describe the data that we use, as well as the experimental setup. Next, we define the different metrics that we use to evaluate our methodology along with a discussion on how to interpret them. Then, we explore the impact of different cloth material models, followed by an ablation study where we analyze the value of each contribution. Later, we compare our methodology with the current state-of-the-art. Finally, we show a novel motion control property that arises from our proposed disentangled architecture.

\subsection{Experimental setup}
\label{sec:exp_setup}

To run our methodology for a given skinned 3D body model we need a dataset of pose sequences. To do so, we gather a few 3D avatars and pose sequences from Mixamo\footnote{https://www.mixamo.com/}. The motions include a few tens of variations for different kind of actions: walking, running, jumping, turning and spinning. Note that some motions contain combinations of these actions. This totals around $450$ motion sequences, containing around $45000$ poses. For each action, we use $5\%$ for validation and $10\%$ for test. Additionally, in order to compare with state-of-the-art methodologies --PBNS~\cite{10.1145/3478513.3480479} and SNUG~\cite{santesteban2022snug}-- we use AMASS dataset \cite{mahmood2019amass} for SMPL model \cite{loper2015smpl}. For fairness, to allow comparison against SNUG public checkpoint, we train PBNS public code and our methodology on the same data. While poses are discrete in time, we implement a continuous sampling by using Slerp \cite{shoemake1985animating}. This allows training our methodology with arbitrary time steps $\Delta t$. During training, samples $\boldsymbol\Theta$ are batched. For each sequence $\boldsymbol\Theta_t$, we predict the garment for the last $3$ time instants to compute $\mathcal{L}_\text{inertia}$. As explained, it is crucial to back-propagate the inertia loss only through $\mathbf{x}_t$. While the static loss terms can be \textit{safely} applied to all $3$ predictions, we observe this creates a sampling bias that hurts performance.

Balancing terms of each loss are related to desired fabric properties. For the cloth term, the values will depend on the chosen cloth material model. Usually within the range $[5, 15]$ for structural stiffness and $[0, 1]$ for shearing. For bending, we use values in the range $[1\mathrm{e}{-5}, 1\mathrm{e}{-4}]$. For collisions, we set $k_c$ to a value similar to the chosen structural stiffness and a threshold $\epsilon = 4mm$. Higher values for $k_c$ will compromise other metrics without improving generalization. Collision-free predictions will depend mostly on training data distribution. Particle mass $m$ --or mass matrix $\mathbf{M}$-- is computed from vertex area and the chosen fabric surface density. Then, the temporal window will depend on the looseness of the garment. We go from $0.5$ seconds for tighter garments up to $2$ seconds for looser garments. Training times until convergence will differ greatly depending on garment, body and motions. From $1$ hour for simpler garments up to a day for garments that show complex and rich dynamics. Inference is extremely fast, as we show in Tab.~\ref{tab:running_time}. We refer to the supplementary code for additional details. The code will be publicly released.

\subsection{Metrics}
\label{sec:metrics}

Traditionally, specially for supervised approaches, lower values for error metrics indicate better predictions. This is not the case for this specific unsupervised problem. Furthermore, the metrics will behave differently for the static case. For that reason, it must be considered and evaluated separately.

\paragraph{Cloth Model.} This metric must be related to the cloth material model used during training. For mass-spring, edge elongation is the most suitable. For continuous formulations, the strain or triangle deformation is a better choice. This metric will measure in-plane cloth deformations. Cloth is resistant to stretching forces, thus, lower values are desired in this case. Some formulations allow modelling shearing forces separately. Cloth is not resistant to shearing, and thus, lower values do not necessarily imply better predictions.

\paragraph{Bending.} Measured as the average error on dihedral angles w.r.t. rest angles for each pair of adjacent faces. Cloth is not resistant to bending. Lower is not necessarily better. A higher bending stiffness will reduce this error, but generate different wrinkles. Thus, a lower error does not imply better predictions, but a stiffer fabric. Note that a null bending error would only be achieved by the template garment in rest pose. Nonetheless, abnormally high values for this metric might suggest a failed simulation. Ultimately, this property must be assessed qualitatively.

\paragraph{Collisions.} Expressed as the percentage of vertices placed within the body. Collisions are to be avoided. For this metric, lower values are desired. For the dynamic case this metric shows an interesting behaviour in the validation data. First, it rapidly decreases until a minimum. Afterwards, as the network learns to predict cloth dynamics, the metric slightly increases until it plateaus. This behaviour does not appear using a static formulation.

\paragraph{Gravity.} Computed as the potential energy of the predicted garments. This energy is defined relative to an arbitrary $0$. Then, it is not possible to define a \textit{goal} for this metric. Its value will also depend on the garment, fabric density, 3D body and pose data. \textbf{Static case:} the optimal solution is achieved when the garment has reached an equilibrium state. For this case, given the same aforementioned experiment conditions, lower is better. Other metrics must be considered as well. A lower cloth stiffness would allow further stretching in the direction of gravity. In such case, we could not conclude the approach converges to a more optimal solution. On the other hand, we can state that training converged when this metric plateaus. \textbf{Dynamic case:} adding motion to the problem changes the behaviour of this metric. For example, a spinning skirt will raise against gravity when dynamics begin to appear, increasing the value of this metric. Likewise, jumping sequences will make the garment \textit{float} when the falling motion begins. This means the garment is not always at its lowest position. For this reason, it is not possible to conclude that lower values are better. Usually, the static formulation gives lower values for gravity. Therefore, a dynamic model with a lower gravity metric might be due to a lack of cloth dynamics.

\paragraph{Inertia Loss.} Measured as the error between $\mathbf{x}_t$ and $\mathbf{x}^\text{proj}_t$ weighted by the particle mass. As explained in Sec.~\ref{sec:training}, lower values do not translate into true cloth dynamics. We empirically observe it behaves the other way around. As training advances and cloth dynamics are being learnt, the value of this metric increases. On the contrary, under the static formulation, this metric will usually decrease. Similar to how gravity metric indicates model convergence for the static case, this metric does the same for the dynamic case. Convergence in training but divergence in validation shows overfitting. This metric also gives an intuition of the level of motion in the predictions. The value of this metric and its evolution during training will greatly depend on garment, fabric, body and motion data.

It is very important to note that all metrics need to be considered at once. The improvement of a single metric is not sufficient to make conclusions regarding the results.

\subsection{Cloth Model} 

\begin{figure}
  \centering
  \includegraphics[width=\columnwidth]{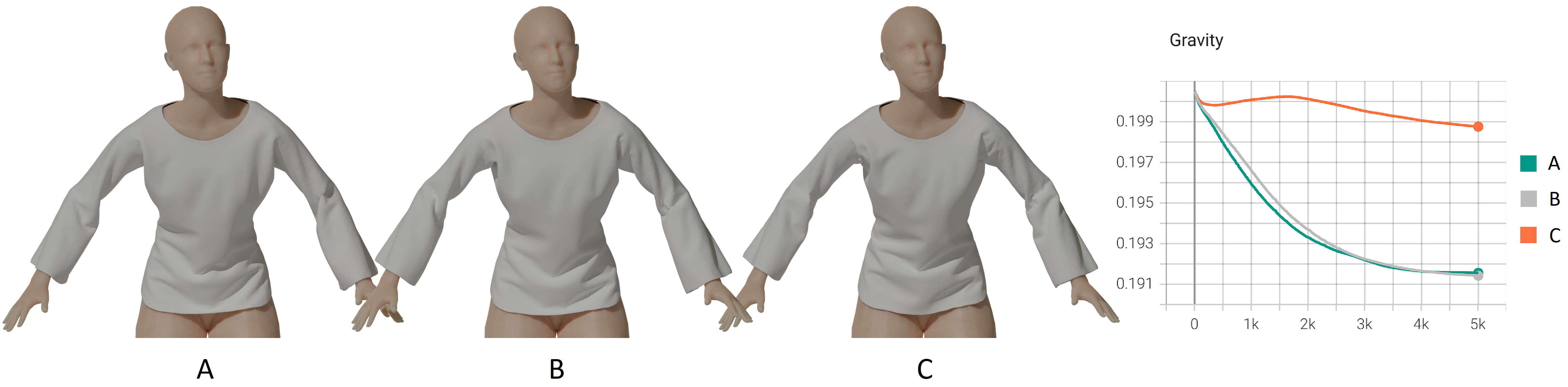}
  \caption{Comparison of cloth material models. Our methodology is compatible with any differentiable formulation for cloth. We test and compare three different ones as a static optimization problem: A) mass-spring model as in \cite{10.1145/3478513.3480479}, B) the continuous formulation proposed by \cite{baraff1998large} and C) the Saint Venant Kirchhoff material used in \cite{santesteban2022snug}. For static optimization, we can use gravity as a measure of convergence.}
  \label{fig:cloth_model}
\end{figure}

We analyze the effect of different cloth material models. To this end, we simulate in a static fashion the same garment, body and poses with different cloth models. We test the mass-spring formulation used by authors of \cite{10.1145/3478513.3480479}, the continuous formulation of \cite{baraff1998large} and the Saint Venant Kirchhoff (StVK) elastic material model as in \cite{santesteban2022snug}. We adapt the material model of \cite{baraff1998large} by squaring their constraints for stretching and shearing. We depict the obtained results in Fig.~\ref{fig:cloth_model}. As shown, the chosen cloth formulation has almost no impact on the qualitative result. Nonetheless, we observe some differences worth mentioning. As opposed to continuous formulations, mass-spring fabric parameters depend on mesh resolution and connectivity. On the other hand, we find that StVK formulation has much more difficulties achieving convergence. We can see this in the gravity plot adjoined to Fig.~\ref{fig:cloth_model}. This formulation may be sub-optimal within the deep learning optimization framework. Finally, the formulation of \cite{baraff1998large} allows explicit control of shearing stiffness. We observed each cloth model scores lower in their respective strain --as expected-- and thus, it is not useful to compare strains quantitatively.

\subsection{Ablation}
\label{sec:ablation}

\begin{figure*}
  \centering
  \includegraphics[width=\textwidth]{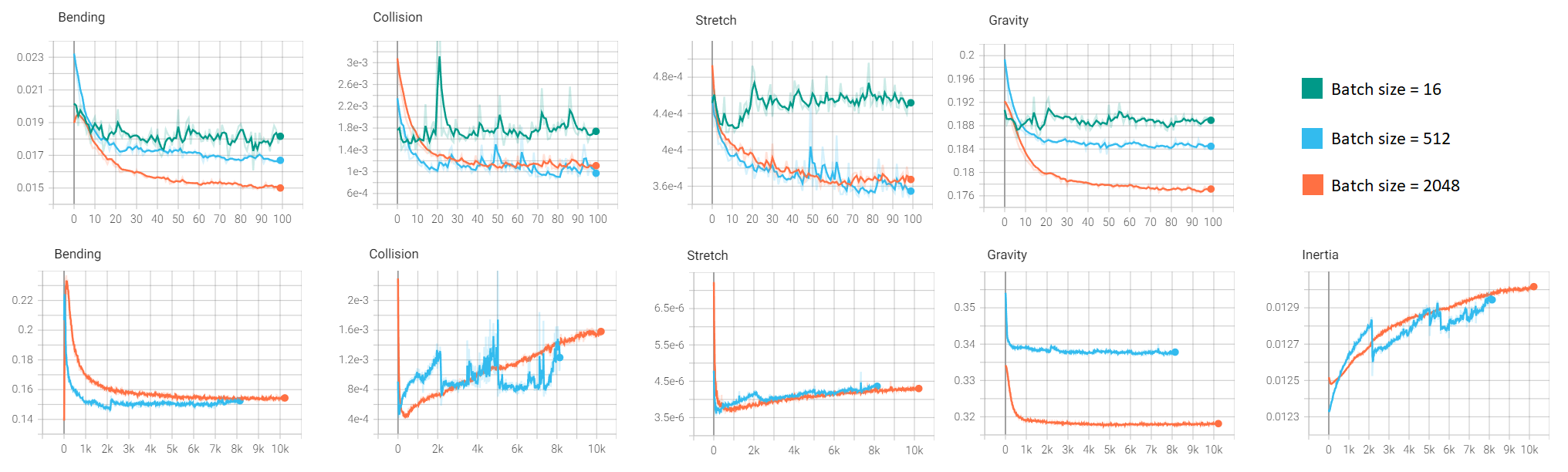}
  \caption{Training progress for different batch sizes. The upper row corresponds to static experiments. The lower row corresponds to dynamic experiments. Using bigger batch sizes significantly increases convergence of the predictions. On the dynamic problem, bigger batches result in a more stable training.}
  \label{fig:batch}
\end{figure*}

\begin{table}[]
  \caption{Static descriptors ablation. We run experiments for each static descriptor under a static formulation. First, raw joint orientation data as quaternions or axis angles. Second, 6D descriptors as in \cite{zhou2019continuity}. Finally, our proposed static descriptors. Our approach shows better generalization and convergence without compromising cloth integrity.}
  \begin{tabular}{lcccc}
  \hline
  \toprule
    & Strain (mm)  & Bending (rads)  & Collision (\%)  & Gravity     \\
  \midrule
  Raw & 0.35463     & 0.0168      & 0.087418     & 0.1845     \\
  6D  & 0.37374     & \textbf{0.01372} & 0.10778     & 0.1811     \\
  6D+G & \textbf{0.35145} & 0.01475     & \textbf{0.073881} & \textbf{0.1765} \\
  \bottomrule
  \end{tabular}
  \label{tab:static_descriptors}
\end{table}

\begin{figure}
  \centering
  \includegraphics[width=\columnwidth]{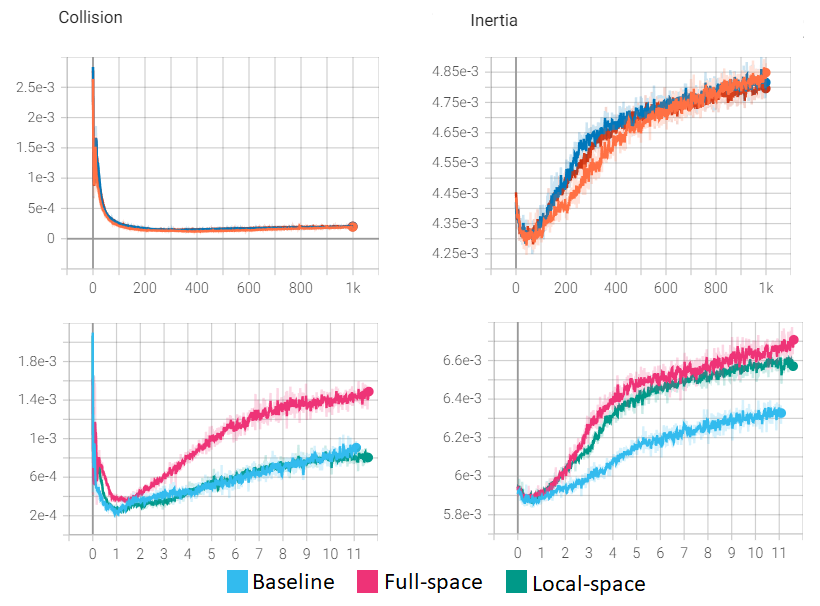}
  \caption{Ablation on dynamic input features. Baseline features contain body pose and root joint velocity. The other experiments contain body pose and joint rotation speeds and accelerations. For the local-space features, we \textit{unpose} accelerations. We evaluate collision (left) and inertia (right) metrics. Upper row corresponds to training data. Lower row to validation data. We see local-space features generalize better --fewer collisions-- while showing a similar level of dynamics as the training data --same inertia curves. We omit the color labels of the training plots (upper row) since they show a very similar behaviour.}
  \label{fig:feats_plots}
\end{figure}

\begin{table}[]
  \caption{Ablation study on network architecture and data augmentation. Experiments with $+$ contain the \textit{improvements} from the previous rows. First row: single encoder. Second row: disentangled encoder. Third row: pose mirroring. Last row: motion augmentation. First and second experiment have no data augmentation. Second, third and fourth experiment have the same architecture. We see how a disentangled encoder and the proposed data augmentations have a beneficial impact. Additionally, we see \textit{gravity} and \textit{inertia} show no significant difference. Therefore, these improvements do not compromise other cloth properties.}
  \begin{tabular}{lcccc}
  \hline
  \toprule
    & Strain & Collisions (\%) & Gravity & Inertia \\
  \midrule
  Entangled   & 9.7840 & 1.101 & 1.068 & 1.889 \\
  Disentangled & 8.2391 & 0.694 & 1.066 & 1.891 \\
  +Mirror    & 8.2146 & 0.505 & 1.067 & 1.864 \\
  +Aug. motions & \textbf{7.6241} & \textbf{0.323} & 1.068 & 1.905 \\
  \bottomrule
  \end{tabular}
  \label{tab:ablation_encoder_augmentation}
\end{table}

\begin{figure}
    \centering
    \includegraphics[width=\columnwidth]{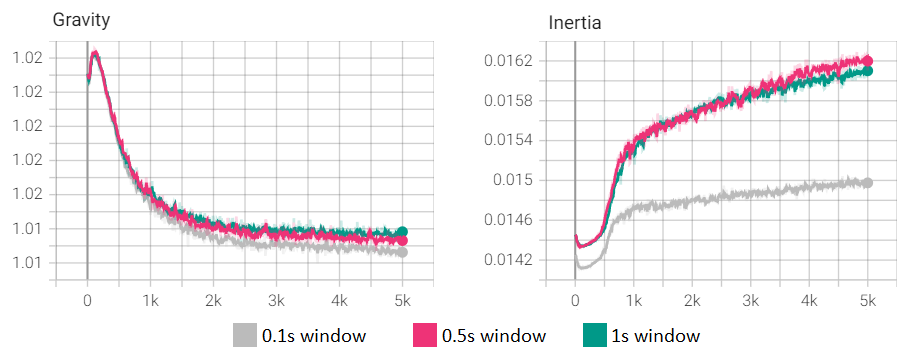}
    \caption{Ablation on temporal window size. We compare neural simulation metrics for a t-shirt with a temporal window size of $0.1$ seconds, $0.5$ seconds and $1$ second. We can see the level of dynamics --\textit{inertia} metric-- is much lower with a small temporal window. We can also observe a slight decrease in \textit{gravity} metric in this case. This is an example of the effect of dynamics on gravity explained in Sec.~\ref{sec:metrics}. On the other hand, we see that further increasing temporal window size has almost no impact in the results. We omit other metrics since their values is almost equal.}
    \label{fig:window}
\end{figure}

\begin{figure}
  \centering
  \includegraphics[width=\columnwidth]{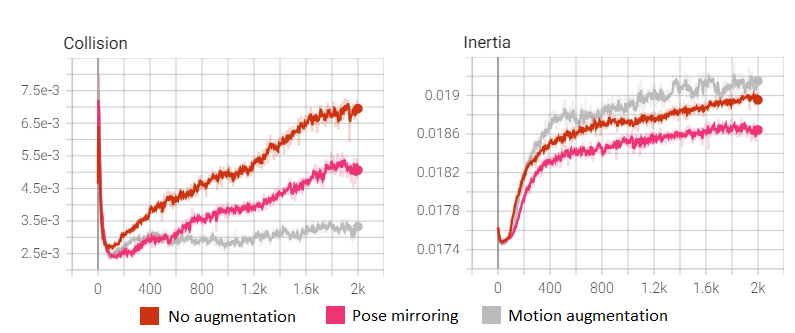}
  \caption{Data augmentation ablation. \textit{Collision} (left) and \textit{inertia} (right) metrics evolution for validation set during training for different augmentation techniques. First: no augmentation. Next: pose mirroring. Finally, a novel motion augmentation technique. The latter is only possible thanks to disentangled encoders. As cloth dynamics are being learnt by the network, collisions slightly increase. Pose mirroring reduces this effect, while motion augmentation mitigates it almost completely. We also see the evolution of cloth dynamics --\textit{inertia} metric-- is not compromised by these changes.}
  \label{fig:augmentation}
\end{figure}

\begin{figure*}
    \centering
    \includegraphics[width=\textwidth]{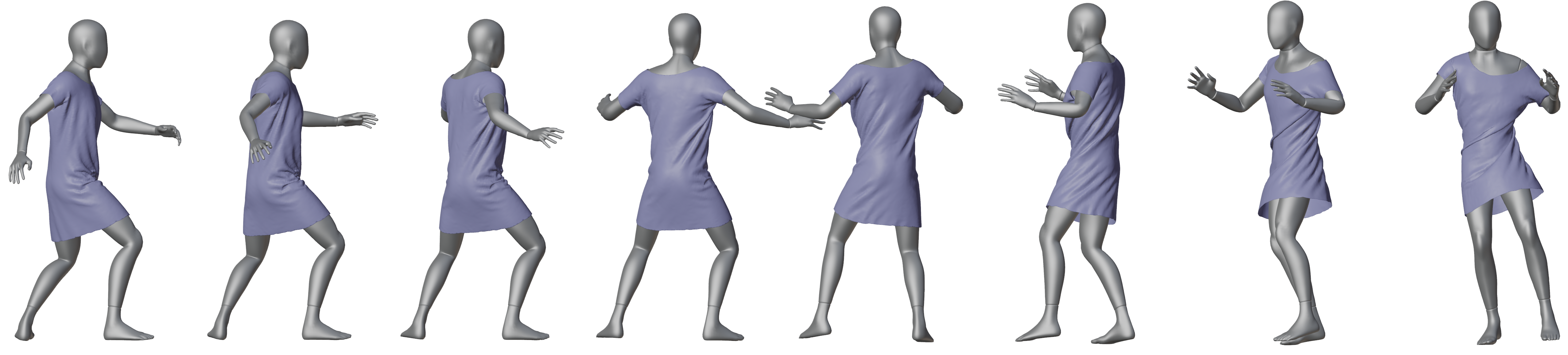}
    \caption{Sample sequence. We illustrate predictions for a dress during a spinning motion. It can be seen how the dress coils up the mannequin body as it spins.}
    \label{fig:sequence}
\end{figure*}

\begin{figure*}
    \centering
    \includegraphics[width=\textwidth]{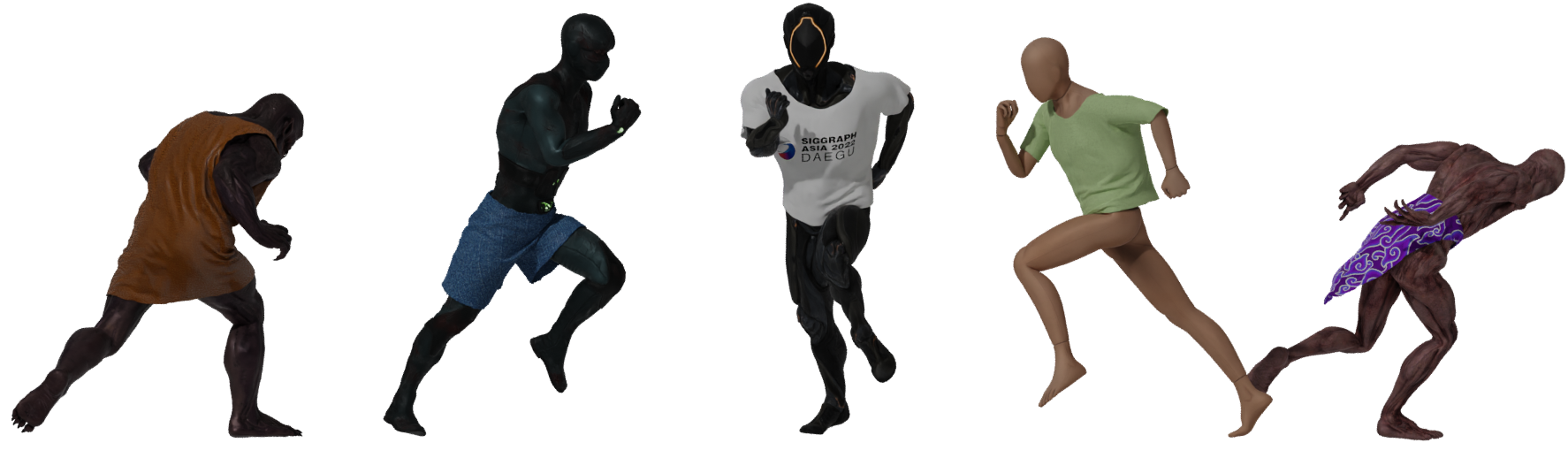}
    \caption{Qualitative samples. Our methodology is compatible with any articulated 3D body and garment. Here we show predictions obtained after neural simulation of different garments draped on different bodies.}
    \label{fig:qualitative}
\end{figure*}

\paragraph{Batch size.} Unsupervised garment animation is quite sensitive to batch size. The reason for this is as follows. Model evolution during training is similar to physical simulation. In the static case, each input sample $\boldsymbol\theta_i$ has a theoretical optimal output $\mathbf{x}_i$. Under supervised training, gradients in the output will point directly towards $\mathbf{x}_i$, with stronger magnitudes for more erroneous predictions. This is not the case for unsupervised garment animation. Gradients will try to greedily update the output cloth by pointing to an intermediate state $\mathbf{\hat{x}}_i$, similar to how a simulation would compute intermediate states until achieving the fully converged solution $\mathbf{x}_i$. Moreover, there is no guarantee that $\mathbf{\hat{x}}_i$ will be closer in space to $\mathbf{x}_i$ than previous predictions. The \textit{path} the model has to follow to convergence is not straight-forward. Furthermore, gradient magnitudes cannot be used as a measure of convergence, except in the extremely unlikely case that their value is $0$ for all the samples in our dataset. On top of that, we have to consider special constraints found in deep learning. Network updates, specially for small batches, can \textit{undo} the work of previous iterations. Then, the network gets stuck in sub-optimal local minima. This produces stiff garments with wrinkling patterns that repeat across different poses. With dynamics, the problem becomes more complex, since the gradients for $\mathbf{x}_t$ depend on $\mathbf{x}_{t-1}$ and $\mathbf{x}_{t-2}$, which are also predictions. Fig.~\ref{fig:batch} shows training metrics for static and dynamic problems with different batch sizes. For the static case, we see that increasing batch size greatly improves model convergence. For the dynamic case, we see that using a small batch size makes the training noisy. There is no plot for batch size $16$ for dynamics since small batches make training unstable and usually fails.

\paragraph{Static descriptors.} We study the impact of the proposed static descriptors under a static formulation. This means training without $\mathcal{L}_\text{inertia}$. We test three different static descriptors. Tab.~\ref{tab:static_descriptors} contains the results obtained. The first row corresponds to raw pose data as quaternions or axis angle representations. Next, 6D descriptors as in \cite{zhou2019continuity}. Finally, the static descriptors proposed in Sec.~\ref{sec:descriptors}. The proposed descriptors present fewer collisions, showing better generalization. They also achieve higher convergence, since gravity is lower. Finally, we see also a lower strain value, which means it did not compromise cloth integrity to minimize the other metrics.

\paragraph{Dynamic descriptors.} We test three different alternatives as motion descriptors. First, as baseline, we use body pose and root joint velocity. This descriptor is the minimum requirement to avoid an ill-posed problem. Second, we gather joint rotation speed and accelerations without \textit{unposing} (full-space). Finally, our proposed descriptors, after \textit{unposing} joint accelerations (local-space). Fig.~\ref{fig:feats_plots} depicts the training plots for collisions and inertia. We omit other metrics since their values barely differ. At left, we have the collision metric. At right, the inertia metric. The first row corresponds to training set plots, and the lower row to the validation set. It is interesting to notice that training plots (top row) are almost the same. 
For validation (bottom row), we observe full-space descriptors show more collisions. These descriptors have a higher variability. Thus, it is a much more challenging task for the network to learn the whole input space. This results in worse generalization. The other two descriptors are local, then, validation set distribution is more likely to fall in the same distribution learnt by the network. For the inertia metric, training and validation plots --although differ in absolute values-- show similar curves. We see the baseline features diverge from the other curves. This indicates a lower generalization. The model has difficulties in extracting meaningful motion information from the baseline representation. Providing of explicit joint rotation speeds and accelerations helps the network to better learn the dependency between body motion and dynamic cloth deformations.

\paragraph{Architecture.} We perform an analysis on model architecture by comparing our disentangled approach against a single encoder. For this experiment, the \textit{entangled} encoder is fed with static and dynamic descriptors. Next, the GRU receives the output encoding. Finally, the decoder predicts garment deformations. The result of this comparison is shown in the first two rows of Tab.~\ref{tab:ablation_encoder_augmentation}. First row corresponds to a single \textit{entangled} encoder. The second row corresponds to the proposed disentangled architecture. We see that separate encoders have better generalization --lower \textit{strain} and \textit{collisions}-- than a single encoder. Additionally, disentangled encoders make training much more stable. Training with an entangled encoder usually fails. This justifies our motivations in the network design.

\paragraph{Temporal window.} As presented in eq.~\ref{eq:well_posed}, predictions need the pose history. We analyze the effect of different temporal window sizes. For this experiment, we neurally simulate a t-shirt with a window size of $0.1$ seconds, $0.5$ seconds and $1$ second. We present the metrics of this experiment in Fig.~\ref{fig:window}. As we can see from the \textit{inertia} metric, a smaller window gives a lower level of dynamics. During training, a reduced input data has a direct impact on the discriminative power of the network. The model cannot properly differentiate motions that are similar within the $0.1$ second window, but have a different past. Because of this, predictions converge to the average of the motions that are close to each other in this reduced input space, which lowers the level of dynamics. Additionally, we observe a lower value for the \textit{gravity} metric for the smaller window size. This is related to the discussion on the effect of dynamics on gravity in Sec.~\ref{sec:metrics}. On the contrary, we observe an even larger temporal window size has no significant impact in cloth dynamics --for this specific case-- but increases the required VRAM and time for training. Therefore, for each specific garment, body and motions, it is important to find a window size that achieves a compromise between dynamics and efficient training. Note that during inference, sequence length can be arbitrarily large. We additionally test the effect of $\Delta t$ by training at different frame rates, $15$ and $60$ (for all other experiments the frame rate is $30$). We notice the model is able to learn realistic cloth dynamics even at lower frame rates. Furthermore, at $15$ fps training is significantly faster. On one hand, the amount of samples is halved. On the other hand, for the same temporal window, the length of the pose sequences is smaller, which means less operations. Finally, the gradients generated by $\mathcal{L}_\text{inertia}$ are larger, and dynamics take less time to appear. On the contrary, increasing the frame rate makes the task much more challenging for the same reasons (but opposed). Nonetheless, training at higher frame rate allows learning finer cloth dynamics. We refer to the supplementary video for a comparison of a motion learnt at different frame rates.

\paragraph{Augmentation.} As the model learns cloth dynamics, collisions in the validation set increase. We study the possibility of mitigating this effect with data augmentation. On one hand, we use standard pose mirroring with probability of $50\%$. On the other hand, leveraging our disentangled approach, we devise a novel motion augmentation technique. To do so, during training, we shuffle the dynamic latent code $\mathbf{z}^D$ for a portion of the samples from each batch ($20\%$ in this experiment). We can apply only the static loss terms to the \textit{augmented} samples. We do not back-propagate gradients to the encoders for these samples. This will give the decoder more data points from $\mathcal{Z}$ and around $\mathcal{Z}^S$, increasing generalization. Tab.~\ref{tab:ablation_encoder_augmentation} shows the effect of the different augmentations. Second row, no augmentation. Third row, pose mirroring. Last row, motion augmentation. We see that for both augmentations, \textit{strain} and \textit{collision} metrics are noticeably reduced, specially for motion augmentation. Moreover, we see it almost completely mitigates the increase of collisions as cloth dynamics are being learnt. This is shown in Fig.~\ref{fig:augmentation}, left plot (\textit{collisions)}. It is also important to notice that \textit{gravity} and \textit{inertia} metrics show very little difference. This means the generalization improvement does not compromise other properties.

We show qualitative results of our methodology for different garments, bodies and motions in Fig.~\ref{fig:teaser}, \ref{fig:sequence} and  \ref{fig:qualitative}. Fig.~\ref{fig:teaser} shows results for SMPL model with different body shapes and the \textit{mannequin} model from Mixamo. The samples show rich and meaningful cloth dynamics. Fig.~\ref{fig:sequence} depicts predictions for a dress during a spinning sequence. As observed, as the motion begins, the dress coils up the mannequin body as we would expect to happen. Finally, in Fig.~\ref{fig:qualitative} we can see how our methodology can generalize to different articulated 3D bodies and garments.

\subsection{State-of-the-art Comparison}
\label{sec:vs_sota}

\begin{figure*}
  \centering
  \includegraphics[width=\textwidth]{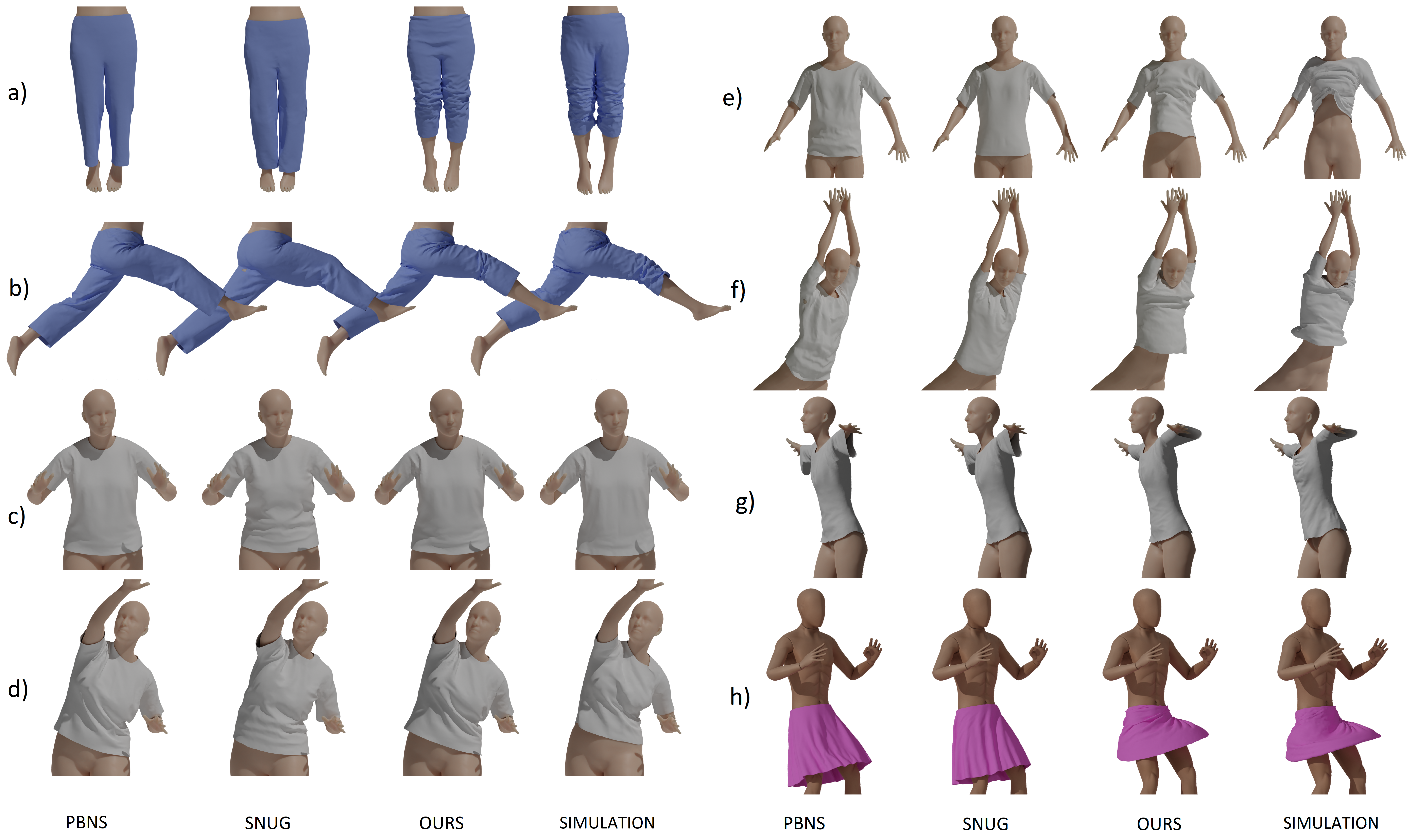}
  \caption{Qualitative comparison for different kind of motions. The motions are: a) jumping, b) leap forward, c) quasi-static pose, d) quasi-static pose, e) jumping, f) dancing jump, g) flapping arm motion and h) fast spin. We use PBNS public code. For samples \textit{a}, \textit{b}, \textit{c} and \textit{d}, we use SNUG public checkpoint. For samples \textit{e}, \textit{f}, \textit{g} and \textit{h}, we implement and train SNUG based on authors public code and paper. Since PBNS uses a static formulation, it shows no cloth dynamics. Then, we see SNUG appears to be unable to show any meaningful dynamics. This is mainly due to an incorrect implementation of the inertia loss. Additionally, looking at the quasi-static samples (\textit{c} and \textit{d}), we notice wrinkling patterns that repeat for most poses. This gives a stiff and less realistic look. Finally, our approach shows dynamic cloth deformations consistent with body motion. We also show results obtained with standard simulation as a reference.}
  \label{fig:vs_sota}
\end{figure*}

\begin{table}[]
  \caption{Quantitative comparison with state-of-the-art: PBNS~\cite{10.1145/3478513.3480479} and SNUG~\cite{santesteban2022snug}. SNUG achieves a much lower value for the inertia term. Nonetheless, see in Fig.~\ref{fig:vs_sota} that their approach shows no cloth dynamics. As explained in Sec.~\ref{sec:metrics}, lower values do not translate to cloth dynamics. }
  \begin{tabular}{lccccc}
  \hline
  \toprule
   & Strain & Bending & Collision (\%) & Gravity & Inertia \\
  \midrule
  PBNS & 3.18    & 0.20     & 0.274    & 0.6274 & 0.937 \\
  SNUG  & 3.89    & 0.20     & 0.283    & 0.6299 & 0.553 \\
  Ours  & 4.2     & 0.15     & 0.276    & 0.6403 & 1.219 \\
  \bottomrule
  \end{tabular}
  \label{tab:vs_sota}
\end{table}

\begin{table}[]
  \caption{Comparison of running times against state-of-the-art: PBNS~\cite{10.1145/3478513.3480479} and SNUG~\cite{santesteban2022snug}. We report performance as \textit{fps}. We propose a standard methodology to measure performance: from raw pose data to final body and garment meshes (for SNUG we report the number in their paper). We test the models with garments of 10K DoF and 250K DoF. Due to its simple formulation and no temporal dimension, PBNS is the fastest approach by far. 
  Our approach, while it does not achieve the efficiency of PBNS, outperforms SNUG. All the approaches run faster than real-time.}
  \begin{tabular}{lccc}
    \toprule
    DoF & PBNS & SNUG & Ours \\
    \midrule
    10K & 7701.5 & 454.5 & 853.9 \\
    250K & 250.5 & - & 206.3 \\
    \bottomrule
  \end{tabular}
  \label{tab:running_time}
\end{table}

\begin{figure*}
  \centering
  \includegraphics[width=\textwidth]{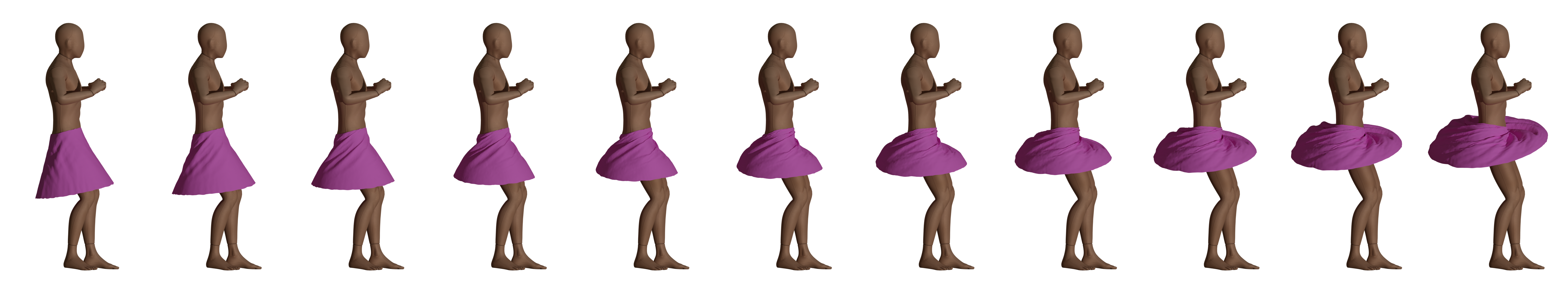}
  \caption{Motion control. In this image we show a still frame of a spinning motion. Thanks to the disentanglement of static and dynamic cloth deformations achieved by the network design, it is possible to control the level of motion in the cloth. To do so, we scale the latent dynamic code to linearly interpolate --and extrapolate-- from the static subspace to the full subspace. Leftmost to middle samples: linear interpolation from static latent code $\mathbf{z}^S$ to latent code $\mathbf{z}$. Middle to rightmost samples: linear extrapolation.}
  \label{fig:motion_control}
\end{figure*}

We evaluate our methodology against recent unsupervised approaches for garment animation. On one hand, a quasi-static solution, PBNS \cite{10.1145/3478513.3480479}. On the other hand, a methodology that claims to model dynamic cloth deformations, SNUG \cite{santesteban2022snug}. SNUG authors provide of a checkpoint we use for comparison. Nonetheless, their methodology is adapted to work with the body shape variability of SMPL. For fairness, we also implement and train their methodology --based on their public code and paper-- using constant body shape. Tab.~\ref{tab:vs_sota} shows quantitative results for each model. Qualitative evaluation is included in Fig.~\ref{fig:vs_sota}. Note we additionally include, as a reference, qualitative results obtained with standard simulation (ArcSim \cite{narain2012adaptive}). We show samples for different body motions: a) jumping, b) leap forward, c) quasi-static pose, d) quasi-static pose, e) jumping, f) dancing jump, g) flapping arm motion and h) fast spin. We use PBNS public code. We know PBNS to be a static solution and therefore, as expected, it does not show dynamic cloth deformations. It is interesting to notice that for quasi-static samples, our solution and PBNS converge to a similar garment. This is the expected behaviour, since our network design ensures the model naturally falls back to a static formulation when there is no input motion. For SNUG, samples \textit{a}, \textit{b}, \textit{c} and \textit{d} are obtained with the official checkpoint provided by its authors. Samples \textit{e}, \textit{f}, \textit{g} and \textit{h} are obtained with our implementation of SNUG. We observe SNUG does not show any meaningful cloth dynamics. After analyzing SNUG, we notice their approach is unable to learn dynamics by design. First, authors observe that $\mathcal{L}_\text{inertia}$ depends on $\mathbf{x}_t$, $\mathbf{x}_{t-1}$ and $\mathbf{x}_{t-2}$ and assume that it is possible to train with sub-sequences of only $3$ body poses. From eq.~\ref{eq:well_posed} we know the whole body motion is needed. While training SNUG, $\mathbf{x}_{t-1}$ and $\mathbf{x}_{t-2}$ are computed from $\{\boldsymbol\theta_{t-1}, \boldsymbol\theta_{t-2}\}$ and $\{\boldsymbol\theta_{t-2}\}$ respectively. This is severely ill-posed, thus, their estimation of $\mathbf{x}^\text{proj}_t$ will be poor. In practice, this means their model will be unable to learn any motion longer than $\sim0.0666$ seconds (3-frame time span for 30 FPS sequences). On the other hand, based on their public code and results, we observe they incorrectly back-propagate $\mathcal{L}_\text{inertia}$ through $\mathbf{x}_{t-1}$ and $\mathbf{x}_{t-2}$. This will give lower values for the loss but not true cloth dynamics. See in Tab.~\ref{tab:vs_sota} how their inertia metric is the lowest by far. This is even more noticeable in looser garments. See the skirt sample \textit{h} under a fast spinning motion in Fig.~\ref{fig:vs_sota}. For circular motions, we have an acceleration $a = \omega^2r$. Back-propagating $\mathcal{L}_\text{inertia}$ through previous frames --as SNUG-- will modify particle locations on all frames to minimize accelerations. Then, their implementation generates gradients pointing inwards that \textit{close} the skirt to minimize $a$ by minimizing $r$. On the other hand, our implementation minimizes the distance between $\mathbf{x}_t$ and $\mathbf{x}^\text{proj}_t$, without modifying the latter. For circular motions, particle velocities are tangential to the circumference. This means that $\mathbf{x}^\text{proj}_t$ will always be outside this circumference. Our approach generates gradients pointing outwards that \textit{open} the skirt to minimize the distance to $\mathbf{x}^\text{proj}_t$. Incorrect back-propagation generates gradients in the \textit{opposite} direction, which makes learning dynamics impossible. This reasoning could be extended to any individual vertex $3$-frame trajectory, as they are locally circular. In their paper, SNUG authors already mention that training on longer sequences resulted in lower inertia values but not true dynamics. Their approach is not learning cloth dynamics, it is smoothing cloth particle velocities (minimizing accelerations). This may give the illusion of a slight cloth dynamic deformation for very specific motions and garments --some jumping motions-- but it will fail for the general case, as we show in all other motions. See supplementary video for additional comparison. Because of these reasons, we cannot consider that SNUG is able to learn cloth dynamics. Additionally, in samples \textit{c} and \textit{d}, we see wrinkling patterns repeating. This happens across most poses, as can be seen in the supplementary video. This gives a stiff and unrealistic look to the cloth. Because SNUG is trained with a small batch size, it suffers from the issue explained in Sec.~\ref{sec:ablation}. Finally, we can see how our approach is able to learn dynamic deformations comparable in quality to standard simulation.

We additionally compare running times of the different methodologies. In the literature we find authors that consider only the forward pass of the network as running time. This is misleading. We propose a new standard for measuring the performance of pose-driven garment animation models. We measure the time it takes from raw pose data (as axis-angle or quaternions) to the final body and garment meshes (for SNUG we report the number in their paper). This gives a much more accurate idea of the running times to be expected on final applications. We report the results in Tab.~\ref{tab:running_time}. We test the models for garments with $10$K DoF and $250$K DoF. PBNS achieves the fastest performance by far due to their simple formulation and no need to model temporal dimension. SNUG authors report in their paper a performance of $454.5$fps. Unfortunately, their runtime GPU code is not open, and so it cannot be adapted to other data sets. According to the authors, it includes collision post-processing by all-pairs testing of cloth and body primitives, parallelized on the GPU.
Finally, our approach does not achieve the performance offered by PBNS, but it still runs significantly faster than real-time. We run all performance tests in a machine with AMD Ryzen 7 5800H and a RTX3060.

We additionally compare the training times of each methodology. PBNS converges in only $10$ minutes. SNUG authors train their approach for $2$ hours. Our training times range from one hour to several hours (up to a day) depending on the complexity of the garment dynamics. Nonetheless, it is important to take into account the complexity of the problem and the solution achieved. PBNS is limited to model static deformations only, which considerably simplifies the problem. SNUG, while it can handle different body shapes using the same network, is trained on a few tens of sequences only, relies heavily on collision post-processing and does not learn cloth dynamics. As seen in the supplementary video, most poses present the same deformations (same wrinkles). Finally, our approach is trained on hundreds of sequences, shows complex dynamics and deformations and requires no post-processing.

\subsection{Cloth Subspace Disentanglement and Motion Control}

The methodology presented in this work is designed to automatically learn disentangled subspaces for static and dynamic cloth deformations. To prove our model is effectively doing so, we linearly interpolate and extrapolate between the static cloth subspace $\mathcal{Z}^S$ and the full cloth subspace $\mathcal{Z}$. To this end, we scale the dynamic latent code given by the dynamic encoder $\mathbf{z}^D$ with $w\in[0, 2]$. That is, $\mathbf{z} = \mathbf{z}^S + w\mathbf{z}^D$. This can be seen in Fig.~\ref{fig:motion_control}. From left to right, the values for $w$ go from $0$ to $2$ with steps of $0.2$. The sample in the middle is the standard network output. This is learnt automatically by the network without any sample or latent code manipulation.

\section{Conclusions and Limitations}
\label{sec:conclusions}

We presented a general framework for unsupervised garment animation. Contrary to previous related works, our work is the first methodology able to learn cloth dynamics without ground truth data. Additionally, we devise a novel disentangled architecture that improves generalization, opens the possibility of a new motion augmentation technique that greatly increases robustness and allows for motion control, which is a useful never seen before property for artists. Also, we provide of detailed analysis and insights on neural cloth simulation that will help future research on the domain. We proved the effectiveness of our methodology with different 3D avatars and garments. Note how the models obtained with our methodology are specific for a given avatar and garment. For new garments or bodies, our methodology needs to be applied again to obtain the corresponding model. Also, while in this work we did not explore body shape generalization, the works of \cite{10.1145/3478513.3480479, santesteban2022snug} proved it. Trained models can generalize to the chosen body parameterization (see Eq.~\ref{eq:integrator}).

Removing the need of gathering ground truth data is a huge advantage in the domain, already noted by previous works \cite{10.1145/3478513.3480479, santesteban2022snug}. Nonetheless, simulation cannot be completely skipped. Instead of being an offline process, simulation and training are now the same. The first time the network sees a given training sample, its corresponding $\mathbf{x}^\text{proj}$ will be an estimation implicitly computed from body motion (transferred to the cloth through garment blend weights). The next epoch, this estimation will be more accurate. So on and so forth. This means that fine cloth dynamics take time to appear, since they have to be indeed simulated within the network. It would not be possible for the network to learn these dynamics without going through these intermediate states. This effect is even greater for looser garments. This means that supervised training will always be much faster, even without considering that unsupervised losses are more computationally expensive. Nonetheless, the methodology still has a significant advantage over supervised approaches, since similar sample motions will contribute to each others neural simulations. On the contrary, even for very similar motions, data gathering through simulation has to be done from scratch for every sequence. Another advantage of unsupervised training is that the network will converge to the simplest solution. That is, similar motions will show similar deformations. On the other hand, offline simulations may show a huge variability for similar motions. Because of this, supervised training needs to deal with noisy data that makes the task much more challenging. Finally, supervised training shows a bias towards lower frequencies and does not satisfy physical constraints without explicit regularization. Then, overall, unsupervised training is still much less time-consuming than data gathering through simulations and will generally converge to simpler, more robust models. We believe an interesting line of research for the future is to study the possibility to kickstart neural simulation with sparse simulated data. Afterwards, fine-tuning through neural simulation will permit efficient training on an arbitrary number of motions with the desired fabric parameters.

Another limitation that neural cloth simulation has yet to address is cloth self-collision. Authors of \cite{10.1145/3478513.3480479} proposed a cloth-to-cloth interaction scheme for different garments or layers of cloth. This formulation is compatible with our methodology. Nonetheless, it is still not enough to model cloth self-collisions. Usually, computer graphics approaches rely on the history of the simulation to prevent self-collisions. That is, an initial state with no self-collisions and careful integration. This works by preventing self-collisions before they happen. This is not possible in deep learning, where history-free self-collision solving methodologies are required. To this end, we consider that adapting the work of \cite{baraff2003untangling} to deep learning is a promising research direction.

\begin{acks}
This work has been partially supported by the Spanish project PID2019-105093GB-I00 (MINECO/FEDER, UE) and CERCA Programme/Generalitat de Catalunya.) This work is partially supported by ICREA under the ICREA Academia programme.
\end{acks}

\bibliographystyle{ACM-Reference-Format}
\bibliography{egbib}

\appendix

\end{document}